\documentclass[]{interact}

\usepackage{epstopdf}
\usepackage[caption=false]{subfig}

\usepackage[numbers,sort&compress]{natbib}
\bibpunct[, ]{[}{]}{,}{n}{,}{,}
\renewcommand\bibfont{\fontsize{10}{12}\selectfont}
\makeatletter
\def\NAT@def@citea{\def\@citea{\NAT@separator}}
\makeatother

\theoremstyle{plain}

\theoremstyle{definition}

\theoremstyle{remark}

\usepackage{multirow}
\usepackage{url}
\usepackage[numbers]{natbib}
\usepackage{enumitem}
\usepackage{diagbox}
\usepackage{subfig}
\newcounter{row}

\newenvironment{imgrows}[1][\columnwidth]%
  {\begin{minipage}{#1}%
   \setcounter{row}{0}%
   \stepcounter{figure}%
  }%
  {\addtocounter{figure}{-1}%
   \end{minipage}%
  }
\newcommand\imgrow
  {\par\noindent
   \refstepcounter{row}%
  
   \raisebox{3.3em}{\makebox[1.5em][r]{(\alph{row})}}
  }

\begin{document}


\title{Predicting and Attending to Damaging Collisions \\for Placing Everyday Objects 
in Photo-Realistic Simulations}

\author{
\name{Aly Magassouba \textsuperscript{a}\thanks{CONTACT Aly Magassouba.
Email: aly.magassouba@nict.go.jp} and Komei Sugiura\textsuperscript{ab} and Angelica Nakayama\textsuperscript{a} and Tsubasa Hirakawa\textsuperscript{c} and Takayoshi Yamashita\textsuperscript{c} and Hironobu Fujiyoshi\textsuperscript{c} and Hisashi Kawai\textsuperscript{a} }
\affil{\textsuperscript{a}National Institute of Information and Communications Technology, 3-5 Hikaridai, Seika, Soraku, Kyoto 619-0289, Japan;\\ \textsuperscript{b}Keio University, 3-14-1 Hiyoshi, Kohoku, Yokohama, Kanagawa 223-8522, Japan;\\ \textsuperscript{c}Chubu University, 1200 Matsumotocho, Kasugai, Aichi 487-8501, Japan} }

\maketitle

\begin{abstract}
 Placing objects is a fundamental task for domestic service robots (DSRs). Thus, inferring the collision-risk before a placing motion is crucial for achieving the requested task. This problem is particularly challenging because it is necessary to predict what happens if an object is placed in a cluttered designated area. We show that a rule-based approach that uses plane detection, to detect free areas, performs poorly. To address this, we develop PonNet, which has multimodal attention branches and a self-attention mechanism to predict damaging collisions, based on RGBD images. Our method can visualize the risk of damaging collisions, which is convenient because it enables the user to understand the risk. For this purpose, we build and publish an original dataset that contains 12,000 photo-realistic images of specific placing areas, with daily life objects, in home environments. The experimental results show that our approach improves accuracy compared with the baseline methods.
\end{abstract}


\section{Introduction
\label{intro}
}
Domestic service robots (DSRs) are promising solutions to the shortage of labor for supporting older adults and disabled people.
Supporting functions that are required from DSRs involve manipulation tasks \cite{smarr2014domestic,billard2019trends}. Placing everyday objects in designated areas is crucial for DSRs \cite{brose2010role} because it is involved in many daily life tasks such as placing a glass on a table and putting a fruit on counter top. Thus, predicting potential collisions in such motions is highly beneficial for many DSR applications; particularly collisions that may put the user or the robot at risk, or damage the environment. 

\begin{figure}[tp]
   \centering
      \includegraphics[width=0.8\columnwidth]{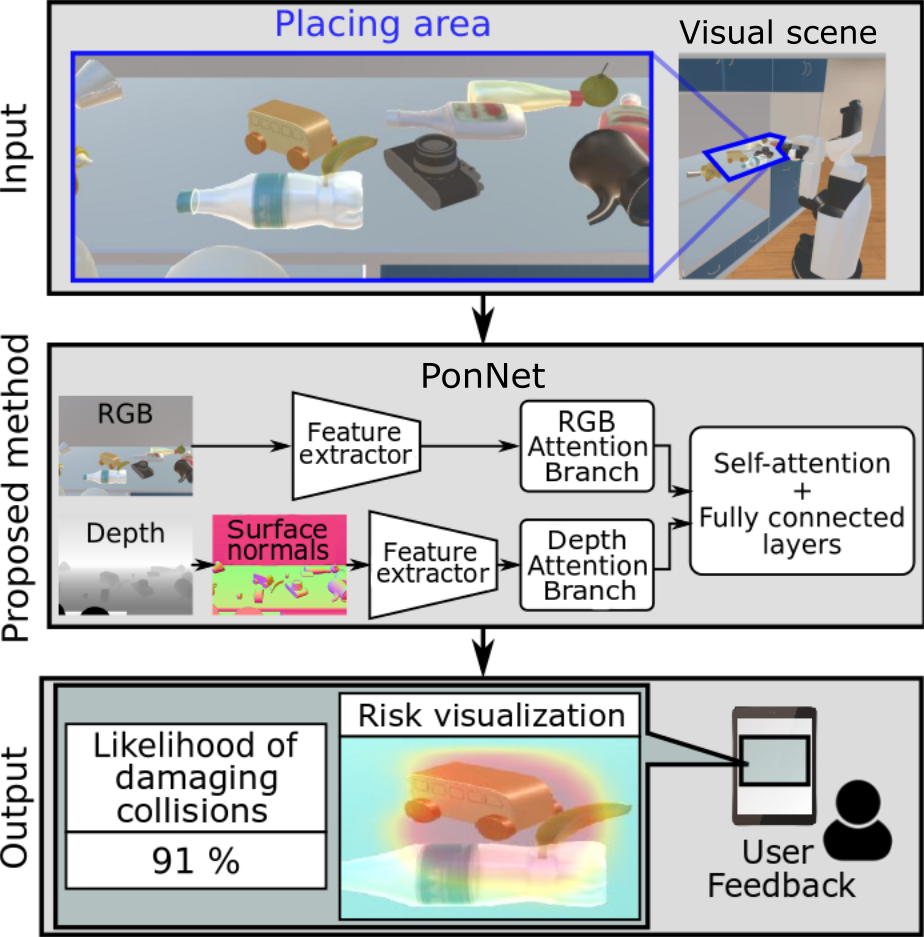}
      \caption{Use case of PonNet: predicting damaging collisions before the placing motion of a DSR.}
   \label{fig:architecture}
 \end{figure}

In this study, we consider the problem of predicting damaging collisions before motion execution. This problem is addressed as a classification task: we infer the likelihood of damaging collisions given a visual scene in which to place an object. A typical use case is shown in Fig. \ref{fig:architecture}.
This problem is particularly challenging because it is necessary to predict what happens if an object is placed in a cluttered designated area. Furthermore, in physical space, not all collisions are damaging. A slight push on an obstacle, or placing an object on a flat obstacle is not damaging. Thus, damaging collisions highly depend on the physics of the objects that are interacting in the environment, which is complex to predict. 

There have been many attempts to evaluate the risk of collisions using data-driven \cite{gualtieri2018pick, magassouba2018multimodal} or rule-based \cite{harada2014validating} approaches.  Although traditional rule-based approaches can partially estimate possible collisions, the estimation generally requires the 3D models of obstacles that are not always available beforehand. On the other hand, they are not required by our method. It can predict collision risk only from camera images. 

To achieve this, we propose PonNet which uses an attention branch network (ABN) \cite{ABN_Fukui}. The ABN is an image classifier, inspired by class activation mapping \cite{zhou2016cvpr} structures, that generates attention maps. The ABN is composed of an attention branch that highlights the most salient portions in an image, given a label to predict.  

In this paper, we extend the ABN by introducing multiple attention branches and an additional self-attention mechanism to predict damaging collisions. Unlike the initial ABN structure, PonNet is a multimodal network that uses RGB and depth inputs: multiple attention branches focus on different parts of these inputs. An additional self-attention mechanism is used to fuse the features extracted from each attention branch. We apply PonNet to the damaging collision prediction (DCP) problem, and compare it with baseline methods. A demonstration video is available at this URL\footnote{\protect\url{https://youtu.be/L-36HZEqFuU}}

The following are our key contributions:
\begin{itemize}
\item We propose PonNet which has multiple attention branches and an additional self-attention mechanism to fuse the output of the attention branches. We explain the method in Section \ref{method}.
\item We build an original dataset BILA-12K with 12,000 designated areas in which to place everyday objects. We explain this dataset in Section \ref{dataset}.
\item We apply PonNet to the DCP problem where a simulated DSR places everyday objects on furniture pieces. We explain the experimental validation in Section \ref{exp}.
\end{itemize}


\section{Related Work
\label{related}
}

Data-driven approaches for manipulation have been widely investigated in robotics \cite{billard2019trends,watanabe2017survey,magassouba2019understanding}. The approach proposed in \cite{jiang2012learning}  trains a classifier to predict the placement suitability of candidate poses based on 3D point-clouds of the object and the environment. The classifier evaluates physical feasibility and stability in addition to human placement preference. The authors of \cite{zeng2018robotic} introduced a 
robotic pick-and-place system that is capable of grasping and recognizing both known and novel objects in cluttered environments. In this work, the robot is endowed with collision avoidance using several motion primitives. Similarly, the authors of \cite{gualtieri2018pick} proposed a system  that uses deep reinforcement learning to address novel objects. The proposed collision avoidance system extracts areas that are free of obstacles.
Most of these works either assume collision-free areas \cite{jiang2012learning} or build specific planning tasks to avoid obstacles \cite{zeng2018robotic, gualtieri2018pick}. However, the latter case provides ad hoc solutions because these planning tasks are adapted to a particular environment configuration, which is why fixed environment setups are used.

A collision prediction method would complement these previous works and improve their generalizability in unknown environments. For this purpose  a collision test algorithm was proposed in \cite{harada2014validating}. This method clusters the working space based on convexity/concavity, contact, and equilibrium properties between the object and environment. However, such an approach requires a deep knowledge of the environment. Thus, the algorithm must be reinitialized for each novel environment.
In our previous work \cite{magassouba2018multimodal}, we addressed the placing task through a classifier, MMC-GAN, to predict the suitability of a designated area from an initial linguistic instruction. Although a collision risk likelihood was implicitly learned by the network, this method was not sufficiently refined to determine the damaging collision risks.

PonNet is a network that predicts and allows the visualization of potential damaging collisions from a visual scene. Damaging collisions depend on the physical properties of the objects interacting in the environment and it was shown in \cite{mottaghi2016happens} that neural networks could learn these properties from visual inputs. PonNet architecture is based on attention branch mechanisms \cite{ABN_Fukui} that allow the visualization of the predicted damaging collisions.  

In contrast to attention mechanisms that are generally integrated into the main network in an unsupervised manner,  attention branches are dedicated branches that generate an attention map from the output labels. An attention branch is designed to focus on specific regions of a feature by generating an attention map. This attention map is then used to improve the performance of the network.
Such an approach was used in our previous works \cite{magassouba2019multi} and \cite{magassouba2020multimodal} in which linguistic and visual modalities were utilized to generate fetching instructions and predict a target to fetch.
Unlike these works, PonNet is based on visual inputs only: RGB and depth modalities are used to predict damaging collisions. Hence, PonNet architecture  can be referred to as an RGBD fusion method. These methods are generally designed for object recognition and detection \cite{chen2018progressively, yang2019adaptive}.

Another contribution of this study is the BILA-12K dataset for DCP. BILA-12K is specifically designed for DSRs because it was collected in simulated home environments (see Section \ref{dataset}). DCP tasks depend on perception and physical reasoning for which PHYRE  \cite{bakhtin2019phyre} and CLEVR \cite{johnson2017clevr} datasets have been proposed respectively. However, these datasets are not suitable for the present study. PHYRE considers only two-dimensional environments whereas CLEVR uses only geometric objects in a single environment: neither furniture pieces nor everyday objects are considered.   
\section{Problem Statement
\label{problem}
}
\begin{figure}[t]
\centering
\subfloat{\includegraphics[width=0.30\columnwidth]{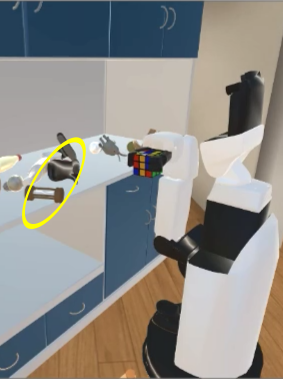}} \hspace{0.01 em}
\subfloat{\includegraphics[width=0.30\columnwidth]{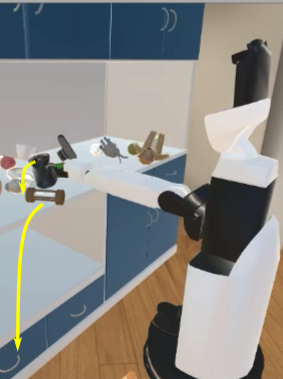}} \hspace{0.01 em}
\subfloat{\includegraphics[width=0.30\columnwidth]{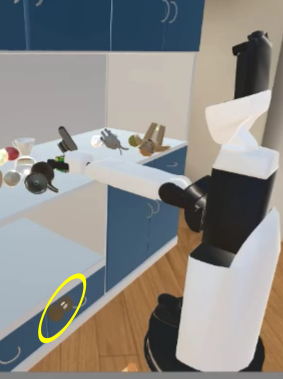}}
\caption{What happens if the robot places the rubik's cube on the shelf? A chain of collision events resulted in a damaging collision. }
\label{fig:placing}
\centering
\end{figure}

\subsection{Task Background}
The typical scene of a placing task, in Fig. \ref{fig:placing}, illustrates the challenges of predicting damaging collisions.
In this example, a fundamental problem is to predict the following: ``What happens if the robot places the Rubik's cube on the shelf?''  
The robot pushed  several objects: a black pot collided with an hourglass and  made it fall from the shelf resulting in a damaging collision. Predicting this chain of events is particularly challenging, because it is necessary to understand the physics of objects and predict the different interactions between them.

Furthermore, in a placing task, several collisions may happen but not all of them can be considered as damaging. In the example illustrated above, slight touches were allowed, however, the collision between the black pot and the hourglass was damaging. In this study, a damaging collision is defined as a collision between two objects, where the absolute relative speed $|\mathbf{v}|$ is greater than a threshold $v_{\text{DC}}$.

These characteristics make it difficult to predict damaging collisions based on classic approaches that use a robot model or free areas. A simple method based on plane detection achieves low prediction accuracy, which is explained in detail in Section \ref{exp}. Furthermore, we aim to visualize the risk of collision as probability, which is convenient because this enables the user to understand the risk and stop dangerous motions in advance. 

\subsection{Task Description: Damaging-Collision Prediction}
Our target task is to predict the likelihood of damaging collisions when DSRs place everyday objects in designated areas. Because the prediction is performed in advance, no video sequence of the placing motion is available. In this paper, we define a DCP problem as follows:
\begin{description}
    \item[Input:] RGBD image of the destination
    \item[Output:] Likelihood of damaging collisions. 
\end{description}
We define the terms used in this study as follows:
\begin{description}
    \item[Target:] The object (e.g., fruits and snacks) that is placed on a destination.
    \item[Destination:] The furniture piece (e.g., shelf) on which the target is placed.
    \item[Obstacle:] An object (e.g., bottle) that is already on the destination. 
\end{description}
We assume that the robot is in front of the destination with the target in its hand. Therefore, the robot can see the destination and reach it. As a result, the placing motion consists of simply extending the robot arm to put down the target. It should be noted that it may be technically possible to introduce force control with a more sophisticated control and planning task. However, solving planning tasks is not our focus. Instead, our study aims at DCP for simple placing strategy. If the likelihood of a damaging collision is above a threshold, then the robot can place the target using a simple strategy. Otherwise, more sophisticated strategies can be chosen, although such strategies may be highly specific to the task environment. For similar reasons, deformable objects are not considered in this study.

We assume that the targets and obstacles are daily life objects. The target candidate should be handled by DSRs, and should be graspable. The  obstacles, targets and destinations are explained more thoroughly in Section \ref{dataset}.
Because the goal of this study is DCP, the evaluation metric is the prediction accuracy for unseen samples.

\subsection{Task Environment}
To make the experimental results reproducible, the simulation environment shown in Fig. \ref{fig:placing} is used. Using a simulation makes the data collection and labeling process faster, whereas with real robots, we must set every scene manually and have a human perform the labeling process. Additionally, a simulated environment is safer for the robot and user: in particular, the evaluation of damaging collisions may be dangerous as the robot collides with its environment.

We use a standardized robot, HSR \cite{yamamoto2018human} with RGBD cameras, as the DSR platform. HSR is emulated through a customized simulator. The simulator extends SIGVerse \cite{inamura2013development} that was used in the World Robot Summit 2018 Partner Robot Challenge/Virtual Space competition. In addition to emulating HSR components following ROS architecture, our simulator is augmented with photo-realistic environments (see Section \ref{dataset}) in configurations similar to those used for DSRs competitions such as RoboCup@Home\cite{I15}. 

\begin{figure*}
\includegraphics[width=1.0\textwidth]{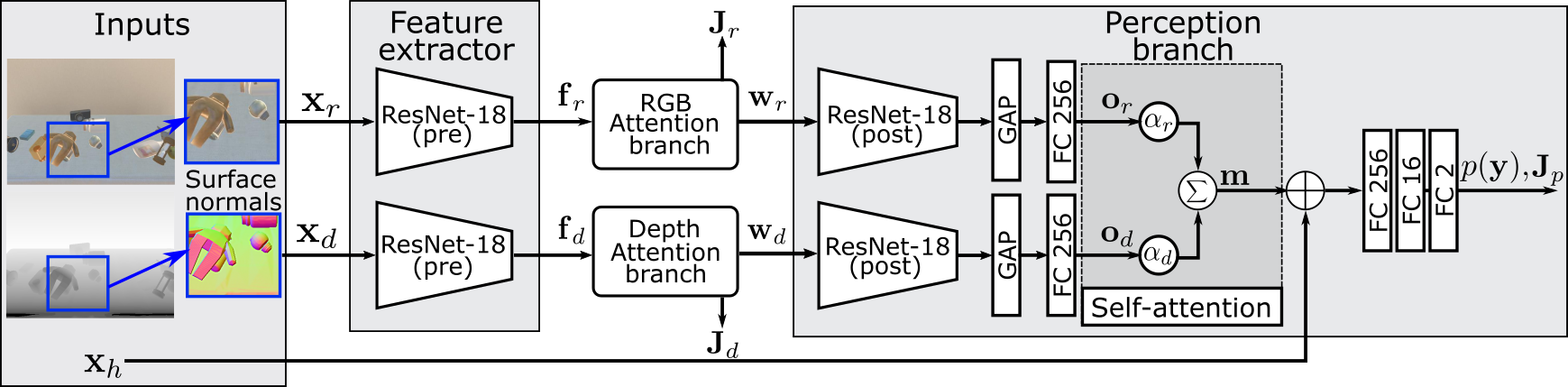}
\caption{Proposed method architecture: PonNet is composed of two attention branches (RGB and depth)  and an additional self-attention mechanism in the perception branch. `ResNet-18 (pre)' represents layers 1-12 of ResNet-18 whereas `ResNet-18~(post)' refers to layers 12-18. Global average pooling (GAP) layers are used for feature dimension reduction in addition to fully connected (FC) layers. }
\label{fig:structure}
\centering
\end{figure*}

\section{Proposed Method
\label{method}
}

\subsection{Novelty}
The PonNet architecture extends the ABN by introducing RGB and depth branches and an additional self-attention mechanism to predict the likelihood of a damaging collision. In this configuration, the RGB and depth modalities complement each other. Indeed, for instance, the depth modality is insensitive to intra-object edges or flat objects unlike the RGB modality but may be noisy. By contrast, transparent objects (e.g., empty bottles) may be detected better using the RGB modality through color variation information.

Thus, PonNet is composed of two attention branches (see Fig. \ref{fig:structure}): RGB and depth attention branches. 
The attention branches  highlight the most salient part of the visual features that are related to the prediction of damaging collision. These features are then input to a perception branch. 
Thereafter, the perception branch fuses the RGB and depth features to predict the likelihood of a damaging collision.
 PonNet has the following characteristics:  
\begin{itemize}
\item PonNet extends the classic RGBD two-stream CNN \cite{chen2018progressively, gupta2014learning} with attention branches.
\item PonNet is composed of multiple attention branches to handle complementary information. The branches are fused by an additional self-attention layer. 
\item Each attention map overlaid on the image input provides a visual explanation of the collision label predicted from each branch.
\end{itemize}

\subsection{Architecture}
\subsubsection{Network Input}
As mentioned above, PonNet takes as input RGB and depth images of the destination, where the robot is expected to place a given target.
Given a destination image $i$, the set of inputs ${\bf x}(i)$ of PonNet is defined as follows:
\begin{equation}\label{eq:input}
    {\bf x}(i)= \{{\bf x}_r(i), {\bf x}_d(i), {\bf x}_h(i) \},
\end{equation}
where ${\bf x}_r(i)$  and ${\bf x}_d(i)$ denote the RGB and depth images, respectively, of the destination area, and ${\bf x}_h(i)$ represents heuristic knowledge about the placing task configuration.  In the following, the index $i$ is omitted for simplicity.  Because we assume that the placing motion is fixed, a placing area is given in advance as a region of interest (ROI) in the destination image. Thus, inputs ${\bf x}_r$  and ${\bf x}_d$ are images of the placing area. Inspired by computer vision literature, the depth input ${\bf x}_d$ is initially transformed into a colorized image based on surface normals \cite{aakerberg2017depth} (see Fig.~\ref{fig:structure}). This method allows us to process a depth image with a similar network architecture as conventional a RGB image. Input ${\bf x}_h$ contains the width, height and length of the target in addition to the camera height. This information is obtained when the robot grasps the target.

\subsubsection{Attention Branches}
\begin{figure}[t]
\includegraphics[width=1.0\columnwidth]{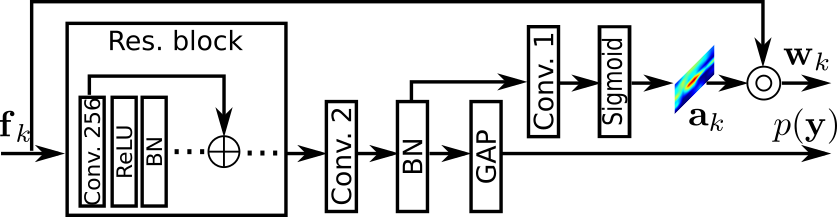}
\caption{Attention branch architecture: An attention map ${\bf a}_k$ is obtained from convolutional (Conv.) layers that are enhanced using batch normalization (BN).  The output  ${\bf w}_k$, with $k \in \{r,d\}$, is obtained from \eqref{equ:w_map} which is represented as `$ \circledcirc $'. `Conv. n' represents convolutional layers of dimensions 14 $\times$ 14 $\times$ n.}

\label{fig:attention}
\centering
\end{figure}

PonNet is composed of two attention branches (see Fig. \ref{fig:attention}) that generate attention maps ${\bf a}_r$ and ${\bf a}_d$ for the RGB and depth modalities, respectively. From these attention maps, the most salient areas of ${\bf x}_r$ and ${\bf x}_d$ are emphasized to predict the likelihood of a damaging collision.

The attention branches are initially input with features maps ${\bf f}_r$  and ${\bf f}_d$  obtained using a backbone CNN feature extractor. We use ResNet-18 as the base network: the feature maps are obtained from the 12$^{th}$ layer. Thereafter, similar to the ABN architecture \cite{ABN_Fukui}, these features maps are processed through residual block layers. In each branch, the output of the residual blocks is processed with a convolutional layer with two kernels, followed by a batch normalization (BN) layer. The first kernel is trained to emphasize features related to damaging collisions whereas the second kernel emphasizes non-damaging collision features. The convolutional layer is then averaged using a global average pooling (GAP) layer from which the likelihood of damaging collisions is predicted. Cross-entropy-based losses $J_{r}$ and $J_{d}$ are minimized by the two attention branches. 

In parallel, an attention map is obtained from an attention block layer. The attention block is composed of a convolutional layer with a single kernel followed by ReLU and sigmoid functions. Weighted feature maps ${\bf w}_r$ and ${\bf w}_d$ are obtained using: 
\begin{equation}\label{equ:w_map}
\begin{aligned}
    {\bf w}_{r}&= (\mathbf{1}+ {\bf a}_{r}) \odot {\bf f}_{r}\\
    {\bf w}_{d}&= (\mathbf{1}+ {\bf a}_{d}) \odot {\bf f}_{d},
    \end{aligned}
\end{equation}
where $\odot$ refers to the Hadamard product. 

\subsubsection{Perception Branch}
In the perception branch, ${\bf w}_{r}$ and ${\bf w}_{d}$ are processed by the backbone CNN ResNet-18 and a GAP for dimension reduction. The features ${\bf o}_{r}$ and ${\bf o}_{d}$ are then obtained from the subsequent fully connected (FC) layer. These features are combined to predict the likelihood of a damaging collision. A classic method consists of concatenating these two features. However, simply concatenating them may limit the accuracy of PonNet, because, for each sample, the depth and RGB images contain different information. Instead, we use a self-attention layer \cite{hori2017attention} which selectively uses the depth and RGB features to improve DCP accuracy. The output ${\bf m}$ of the self-attention layer is given by
\begin{equation}
    {\bf m} = \alpha_r {\bf o}_{r}  + \alpha_d  {\bf o}_{d}, \label{att weight}
\end{equation}
where $\alpha_r$ and $\alpha_d$ are parameters trained by the network. These parameters are computed as follows:
    \begin{equation}\label{balancing}
        \alpha_k  = \frac{\exp{(e_{k})}}{\sum_{j=1}^K \exp{(e_{j})}}, \\
    \end{equation}
and  
 \begin{equation}\label{weights}
        e_{k} = {\bf V}^{T} \tanh{({\bf W}_{k} {\bf o}_{k} + {\bf b}_k)},
\end{equation}
where the index $k \in \{r, d\}$ denotes the given input modality. The weight matrix ${\bf W}_{k}$, ${\bf V}$ and bias vector ${\bf b}_k$ are learnable parameters.

Finally, ${\bf x}_h$, which characterizes the camera height and the size of the target, is input to the network. The likelihood $p({\bf y})$ of a damaging collision  is then obtained by processing  the concatenation of ${\bf m}$ and ${\bf x}_h$ in FC layers. In the same manner as the attention branches, the cross-entropy loss $J_{p}$ is minimized by the perception branch.

\subsubsection{Network Loss}
To predict the likelihood $p({\bf y})$ of a damaging collision, the loss function $J_{total}$ of the network is given by 
\begin{equation}\label{equ:J_p}
J_{total} = \lambda_r J_{r} + \lambda_d J_{d} + \lambda_p J_{p},
\end{equation}
where $\lambda_r$, $\lambda_d$ and $\lambda_p$ denote the different loss weights.
Given $J$ as a generic notation for $J_{r}$, $J_{d}$ and $ J_{p}$, the cross-entropy loss is expressed as follows
\begin{align} \label{equ:J}
    J &= -\sum_n \sum_{m} y^{*}_{nm} \log p(y_{nm}),
\end{align}
where $y^{*}_{nm}$ denotes the label given to the $m$-th dimension of the $n$-th sample, and $y_{nm}$ denotes its prediction.    
The attention and perception branches predict the same labels.


\section{Dataset: BILA-12K
\label{dataset}
}
\begin{figure}[!h]
\subfloat{\includegraphics[width=0.489\columnwidth]{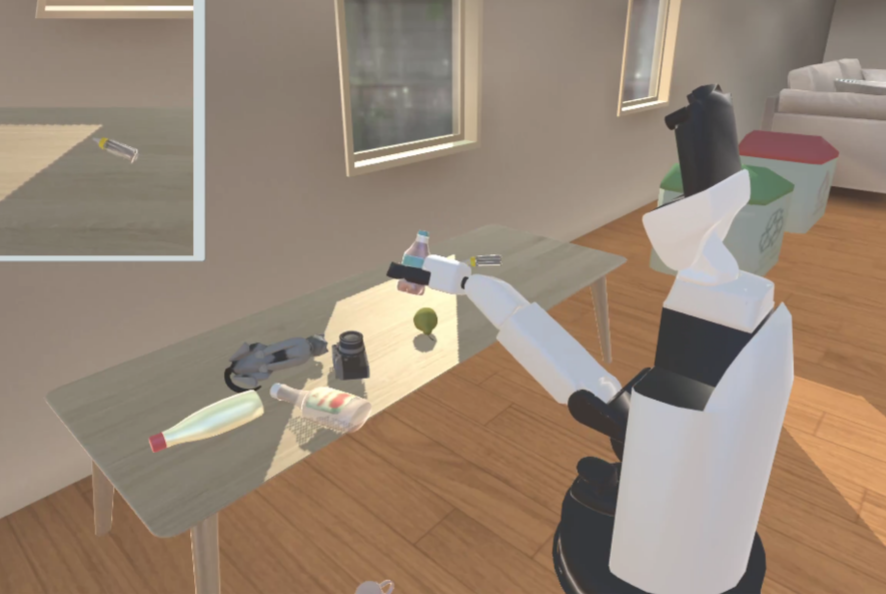}} \enskip
\subfloat{\includegraphics[width=0.489\columnwidth]{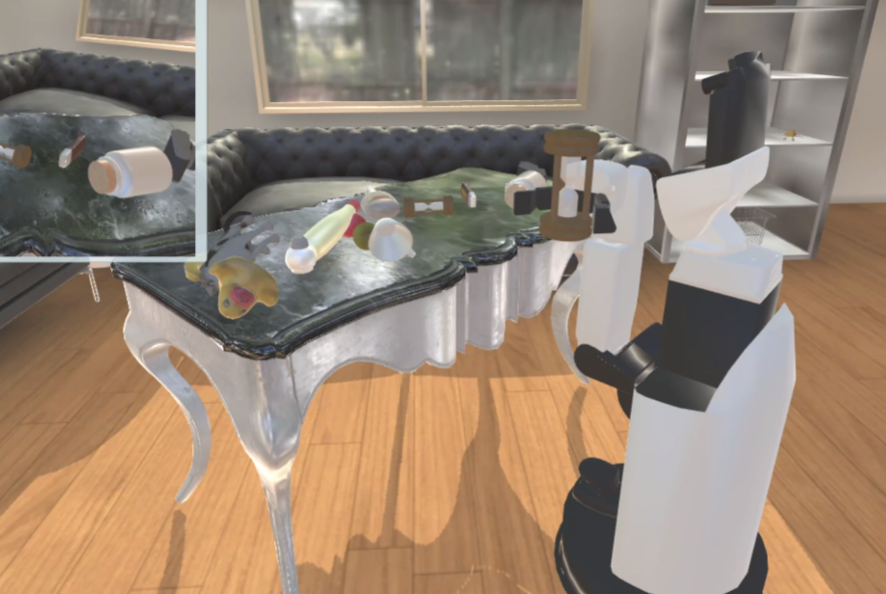}} \\\vskip -0.1pt
\subfloat{\includegraphics[width=1\columnwidth]{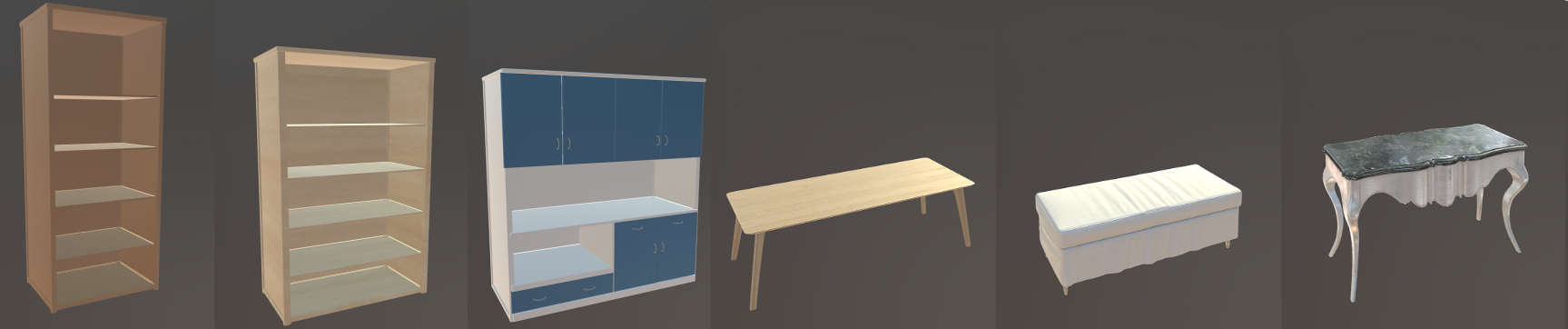}}\\\vskip -0.1pt
\subfloat{\includegraphics[width=1\columnwidth]{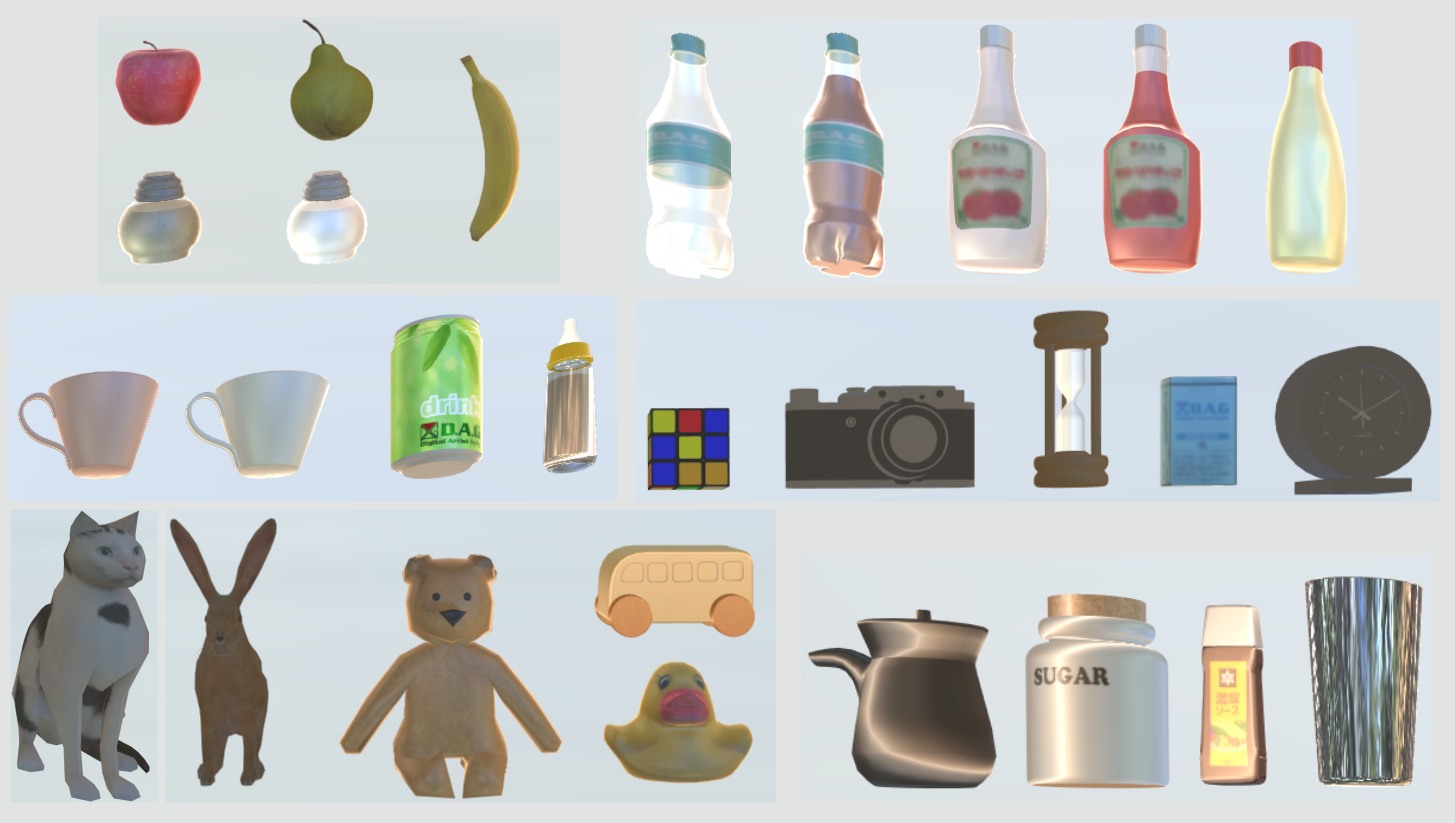}}
\caption{Simulator environment setup. Top-row figures show the placing motion in different conditions. Middle and bottom-row figures illustrate samples of the destinations and obstacles/targets.}
\label{fig:env_setup}

\centering
\end{figure}

\subsection{Dataset Construction Overview}
The top row figures in Fig. \ref{fig:env_setup} show the simulation environment used in data collection. 
The dataset BILA-12K (available at this URL\footnote{\protect\url{https://komeisugiura.jp/software/bila12k/latest.tgz}}) was collected from a simulated home environment, as described in Section \ref{problem}. Simulated environments allow the automatic collection of large-scale data in various configurations. BILA-12K can be described as follows: 
\begin{itemize}
\item BILA-12K was collected from photo-realistic environments based on Unity engine. Each environment was procedurally generated with everyday objects and furniture as illustrated in Fig. \ref{fig:env_setup}.
\item BILA-12K is composed of 12,000 RGBD samples of designated areas in which a target can be placed.
\item BILA-12K contains the collision statistics for all samples through the placing motion performed by a standardized robot, HSR. A video of the placing motion was also recorded for further analysis and verification.
\end{itemize}
The procedure given below describes the steps used to build this dataset.


\subsection{Procedure}
\begin{enumerate}[label=\arabic*)]
	\item{\bf Environment Setup:}\label{proc:one}
As shown in Fig. \ref{fig:env_setup}, the simulator procedurally generated home environment. For more variety, we considered 12 distinct locations, with various background and lighting conditions, distributed among different rooms. BILA-12K is composed of 12,000 samples, that is, 1,000 samples for each location.
	\item{\bf Destination:}
For each sample, HSR's position was initialized in front of a randomly selected destination. A destination corresponds to everyday furniture such as tables, shelves and sofas. Each destination has a different shape, material and reflection properties. A set of six destination candidates was considered. It should be noted, that destinations such as shelves have multiple levels that could be selected as placing areas.
	\item{\bf Obstacles:} 
 The obstacles refer to daily life objects in the environment. From a pool of 28 objects, a random number of obstacles were randomly scattered on the destination. To increase the variation of obstacle positions and orientations, all obstacles were dropped from a given height.
\item{\bf Target:}
Similar to obstacles, a target was selected from a set of 15 daily objects that are graspable by HSR.
\item{\bf Placing Motion:}
HSR recorded through its RGBD camera the state of the destination as a 640 $\times$ 480 image, in addition to the camera height and target dimension. Thereafter, the robot could perform the placing motion (please see the accompanying video) by extending its arm. The collision statistics were then recorded for each task. As our focus is on DCP, no placing strategy nor obstacle avoidance were implemented. Similarly, the simulator did not handle slip or friction. These parameters are out of the scope of this study.
\item{\bf Repeat:}
The procedure was repeated from step 1) until a sufficient number of samples was collected.
\end{enumerate}

\subsection{Labeling Strategy}
Each placing motion was then automatically labeled by the simulator. Two labels, DC (damaging collision) and NDC (no damaging collision), were used for the DCP task.
The damaging collision speed threshold was defined as $v_{\text{DC}}$~=~$0.1$~$\text{m/s}$. 
If collisions above this threshold were detected, the sample was labeled DC. Otherwise the sample was labeled NDC. To evaluate PonNet,
BILA-12K was randomly split into training (10,000 samples), validation (1,000 samples) and test (1,000 samples) sets. After removing invalid samples, we  obtained the statistics summarized in Table \ref{tab:dataset}.

\begin{table}[t]
\caption{ Statistics for the PonNet dataset where two labels DC (damaging collision) and NDC (no damaging collision) are used to characterize collisions}
\label{tab:dataset}
\centering
\begin{tabular}{c|ccc|c}
\hline
$\#$ & Train& Valid.& Test& $\sum$([\%])\\
\hline
DC&4807&492&448& 5747 (48.7) \\

NDC&5074&496&490& 6060 (51.3)\\
\hline
$\sum$([\%])& 9881 (83.7) & 988 (8.4)& 938 (7.9)&11807 (100)\\

\hline
\end{tabular}

\end{table}





\section{Experiments
\label{exp}
}

\subsection{Experimental Setup}

The input images were rescaled to 224$\times$224.
Each of them contains the region where the robot hand is placing the target, and was extracted from the camera image.

The CNN backbone of PonNet was based on the ResNet-18 network  as mentioned in the method section. It was initialized with ImageNet \cite{russakovsky2015imagenet} weights and fine-tuned on BILA-12K. 
The feature maps that were input to the two attention branches were extracted from the 12$^\text{th}$ layer of ResNet-18. Each feature map, ${\bf f}_r$ and  ${\bf f}_d$, had the dimensions 14$\times$14$\times$256. 
The attention branches were composed of three residual blocks with full pre-activation \cite{he2016identity}. To comply with the feature map dimensions, the convolutional layers in the attention branches had the dimensions 14$\times$14. The second CNN corresponded to the remaining lower layers of ResNet-18. More specifically, the processed features of dimensions 14$\times$14$\times$256 were input to the network (12$^\text{th}$ layer), and the features of dimensions 7$\times$7$\times$512 were output from the 18$^\text{th}$ layer. From a GAP and a flattening process, a feature vector of dimension 512 was obtained.
The attention layer used weights of dimensions 256$\times$256. The self-attention block is followed by FC layers of size 256, 16 and 2. 

The loss weights were empirically set to 1 for the attention branches and $0.3$ for the perception branch. PonNet had approximately 18 M parameters and was trained on a machine equipped with a Tesla V100 with 32 GB of GPU memory, 768 GB RAM and an Intel Xeon 2.10 GHz processor. The results were reported after 150 epochs when the training loss was reduced by approximately 98\%. With this setup, it took around one and a half hours to train PonNet with a batch size of 48 samples and a learning rate of 5$\times$10$^{-4}$. The parameters of PonNet are summarized in Table \ref{tab:param}. 

\begin{table}[b]
\caption{ Parameter settings and structures of PonNet}\label{tab:param}
\centering
\begin{tabular}{|c|l|}
\hline
PonNet & Adam (Learning rate= $5e^{-4}$, \\
 Opt. method  &$\beta_1=0.99$, $\beta_2=0.9$) \\
\hline
Network input size & $[224\times224]$\\
\hline
Backbone CNN  & ResNet-18\\
\hline
Res. Block  & full pre-activation\\
\hline
 & Input: $[14\times14\times256]$  \\
 \cline{2-2}
 Attention  & Conv.layer : $[14\times14\times2]$  \\
\cline{2-2}
 Branch & Att. layer: $[14\times14\times1]$  \\
 \cline{2-2}
 & Output:   $[14\times14\times256]$ \\
\hline
Perception& Att. layer : 256 \\
 \cline{2-2}
Branch& FC: 256, 16, 2 \\
\hline
Weight   &$\lambda_d=1, \lambda_r=1$,  $\lambda_p=0.3$  \\
\hline
 Batch size & 48  \\
 \hline
\end{tabular}
\end{table}

\subsection{Quantitative Results}
We compared PonNet with a baseline method: classic RGBD plane detection algorithm \cite{wang2018plane} (available at this URL\footnote{\protect\url{https://github.com/chaowang15/RGBDPlaneDetection}}) for free area detection. We also performed several ablation studies with  different PonNet types using no attention, i.e., ResNet-50 only (type1) and ResNet-18 only (type2) and a single attention branch \cite{ABN_Fukui} (type3). Because these methods are inherently monomodal approaches, RGB only, depth only and RGBD inputs were tested. It should be noted that for the RGBD configuration, RGB and depth inputs were concatenated channel-wise, through an early fusion scheme. 
Additionally a variant of PonNet without self-attention (type4) in the perception branch was also tested. Except for the plane detection methods, all of the above methods have exactly the same inputs, i.e., RGB and colorized depth images.

All the above methods were evaluated for test set accuracy when the best validation accuracy was obtained. Hence, these results reflect more objectively the accuracy of each method on unseen data. The quantitative results of all methods are summarized in Table \ref{tab:results} which reports the average accuracy and  standard deviation for five trials. 
In the table, the ``BB'' column shows the backbone network as either of ResNet-50 (``50'') or ResNet-18 (``18''). 
The ``AB'' column shows the type of attention branch as either of single ABN (``S'') or multiple ABN (``M'').
The ``SA'' column shows whether self attention is introduced or not.

\begin{table}[t]
\caption{Average test set accuracy and standard deviation over five trials for DCP.  
The ``BB'' column shows the backbone network as either of ResNet-50 (``50'') or ResNet-18 (``18''). 
The ``AB'' column shows the type of attention branch as either of single ABN (``S'') or multiple ABN (``M'').
The ``SA'' column shows whether self-attention is introduced or not.}

\label{tab:results}
\centering
\begin{tabular}{l|c|c|c|c|c}
\hline
\multicolumn{1}{l|}{\bf Method} & BB & AB & SA &\multicolumn{1}{c|}{\bf Input} & \multicolumn{1}{c}{\bf Accuracy}  \\
\hline
\hline
 Plane detect. & $-$ & $-$ & $-$ & RGBD & 82.5\\
 \hspace{2mm}(baseline) & & & & & \\
\hline
\multicolumn{1}{c|}{} &&&&RGB & 88.25$\pm$0.45\\
PonNet-type1  & 50 & $-$ & $-$ & D &87.89$\pm$0.88\\
\multicolumn{1}{c|}{} &&&& RGBD & 88.58$\pm$0.67\\
\hline
 \multicolumn{1}{c|}{} &&&& RGB &88.82$\pm$0.68\\
PonNet-type2 & 18 & $-$ & $-$ & D &88.42$\pm$0.22\\ \multicolumn{1}{c|}{}&&&&RGBD& 89.25$\pm$0.33\\
\hline
\multicolumn{1}{c|}{} &&&& RGB &89.85$\pm$0.45\\
PonNet-type3  & 18 & S & $-$ & D &89.51$\pm$0.82\\
\multicolumn{1}{c|}{}  &&&& RGBD&90.18$\pm$0.11\\
\hline
PonNet-type4 & 18 & M & N & RGBD & 90.47$\pm$0.32\\
\hline
PonNet (ours) & 18 & M & Y & RGBD &{\bf90.94$\pm$0.22}\\
\hline
\end{tabular} 

\end{table}

The best accuracy was obtained by the proposed architecture, PonNet, with an average of 90.94$\%$ of correct predictions. Our method outperformed the baseline plane detection method by $8.4\%$. Because the plane detection method did not have variation in output, the standard deviation is not shown. Furthermore, the ablation study emphasized the contribution of each part of the network. We observed that the attention branch mechanism improved the prediction accuracy compared to the PonNet-type1 and PonNet-type2. Furthermore, PonNet-type2 performed better than PonNet-type1: we hypothesize that the number of layers of ResNet-18 was more adequate for our dataset. Using several branches instead of a single one is also relevant as the accuracy increased. Similarly, the self-attention mechanism improved PonNet accuracy from 90.47\% to 90.94\%. 

\begin{figure}[!b]
\begin{imgrows}%
\imgrow\label{fig:one}
  \includegraphics[width=0.3\columnwidth]{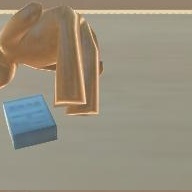}
  \includegraphics[width=0.3\columnwidth]{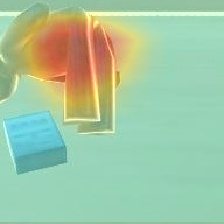}
  \includegraphics[width=0.3\columnwidth]{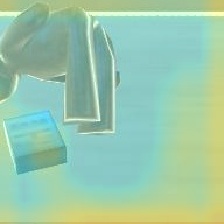} \vskip 1pt
\imgrow\label{fig:two}
  \includegraphics[width=0.3\columnwidth]{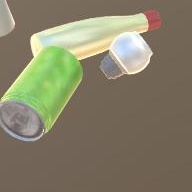}
  \includegraphics[width=0.3\columnwidth]{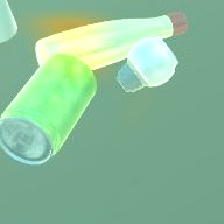}
  \includegraphics[width=0.3\columnwidth]{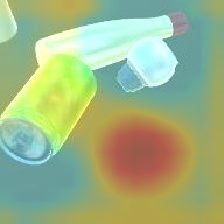} \vskip 1pt
\imgrow\label{fig:three}
  \includegraphics[width=0.3\columnwidth]{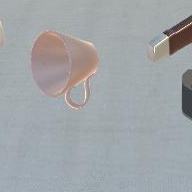}
  \includegraphics[width=0.3\columnwidth]{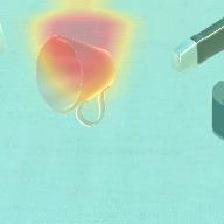}
  \includegraphics[width=0.3\columnwidth]{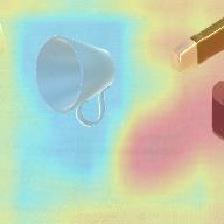} \vskip 1pt
\imgrow\label{fig:four}
  \includegraphics[width=0.3\columnwidth]{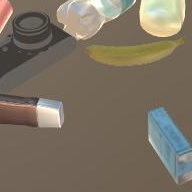}
  \includegraphics[width=0.3\columnwidth]{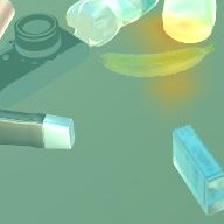}
  \includegraphics[width=0.3\columnwidth]{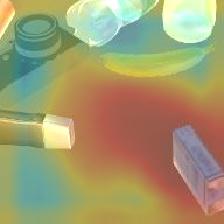}
\end{imgrows}

\caption{Qualitative results of DCP using PonNet.  The first column is the input image whereas the second and third columns are the RGB and depth attention maps. (a) True positive, (b) True negative, (c) False positive, (d) False negative.}

\label{fig:qualitative}
\end{figure}

Additionally, as expected, our ablation study demonstrated that multimodal approaches outperformed monomodal approaches regardless of the network tested. However, RGBD fusion through PonNet architecture was more efficient. This result emphasizes that RGB and depth modalities complement each other, as claimed initially.

\subsection{Qualitative Results}


PonNet architecture embeds attention branches that allow visual explanation. Such a feature is particularly desirable for analyzing the functioning of neural networks that are typically considered as black boxes. Fig.~\ref{fig:qualitative} illustrates the attention of PonNet output by the RGB and depth attention branches for a given input. Figs.~\ref{fig:qualitative}(a) and (b) show accurate predictions, whereas Figs.~\ref{fig:qualitative}(c) and (d) illustrate the failure cases.  The attention maps of the RGB and depth modalities are given for each case.
These results emphasize a key point of PonNet: the visualization of the predicted damaging collisions, where the most salient areas of a given image are in red. The attention maps for the RGB and depth images were particularly different and can be interpreted in the following manner:
\begin{itemize}
    \item RGB modality attends to the obstacles that may lead to damaging collisions. As illustrated in Fig.~\ref{fig:qualitative}(a), small and flat obstacles (e.g., cigarette) are not attended unlike the teddy bear that lead to a damaging collision.
    \item Depth modality attends to the areas in which the target can be placed safely. In Fig.~\ref{fig:qualitative}(b) there is no risk of collision in the area attended: as observed during the placing task, a slight touch made the soda can roll gently. By contrast, no area is attended in Fig. \ref{fig:qualitative}(a) because the target cannot be placed safely.
\end{itemize}
Interestingly, in addition to giving understandable feedback to users, these attention maps may be used to predict the most suitable areas for the placing task given several candidate areas in the same manner as that in \cite{magassouba2018multimodal}. In the same way, a future study may be about the evaluation  of the attended areas  with a network structure where the target shape is input to the attention branches.

Regarding erroneous predictions, in Fig.~\ref{fig:qualitative}(c) attention provided contradictory outputs: from the depth there was a low likelihood of damaging collision whereas it was the opposite for RGB that focused of the cup. The PonNet self-attention mechanism favored the RGB modality in this case and erroneously predicted a damaging collision. In Fig.~\ref{fig:qualitative}(d) both modalities failed to predict the damaging collision with the cigarette box (in blue): unlike the case illustrated in Fig.~\ref{fig:qualitative}(a), this obstacle was standing on its edge and lead to a damaging collision when knocked down.
This emphasizes the challenge of DCP tasks: the same object in a slightly different position may lead to two different outcomes.

\subsection{Error Analysis}
PonNet erroneous predictions were also quantitatively analyzed. The confusion matrices reported in Table \ref{tab:conf},  show the classification results  for PonNet in addition to the detailed predictions of the RGB and depth attention branches.

By analyzing the PonNet results, we observed that most erroneous cases were false-positive predictions: a damaging collision was predicted whereas this is not actually the case. These errors made the system more conservative about the collision risk for the placing motion. By contrast, there were only 32 false-negative predictions. These errors are more crucial in a placing task. Despite this, they represented only 3.5\% of all predicted samples, which is reasonable considering the complexity of the DCP task. Regarding the RGB and depth results, they show that the attention branches alone performed differently and with a lower accuracy. Despite this, false-negative predictions were greatly reduced (-20\%) by fusing both modalities as performed in PonNet.

\begin{table}[t]
\normalsize
\caption{Confusion matrices of PonNet, RGB and depth attention  branches. ${\bf y}$ and $\widehat{\bf y}$ represent the ground truth and prediction, respectively.}\label{tab:conf}
\centering
\begin{tabular}{c|cc|cc|cc}
\hline
  \multirow{2}{*}{\diagbox{${\bf y}$}{$\widehat{\bf y}$}}&\multicolumn{2}{c|}{\bf Total}&\multicolumn{2}{c|}{\bf RGB Att.}&\multicolumn{2}{c}{\bf Depth Att.}\\
\cline{2-7}
&DC &NDC &DC &NDC &DC &NDC \\
\hline
 DC & 458& 32 & 450& 40 & 450& 40\\
 NDC  & 60& 388& 72& 376& 64& 384\\
\hline
\end{tabular}
\end{table} 
\subsection{Generalization ability}
\begin{figure}[t]
\begin{imgrows}%
\imgrow\label{fig:tea}
  \includegraphics[width=0.3\columnwidth]{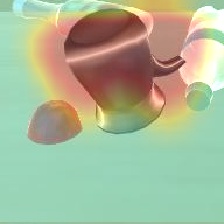}
  \includegraphics[width=0.3\columnwidth]{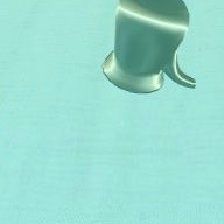}
  \includegraphics[width=0.3\columnwidth]{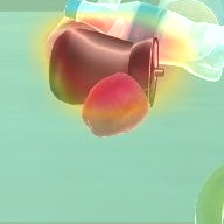} \vskip 1pt
\imgrow\label{fig:cup}
  \includegraphics[width=0.3\columnwidth]{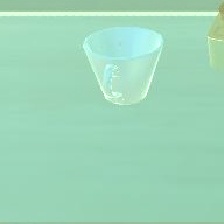}
  \includegraphics[width=0.3\columnwidth]{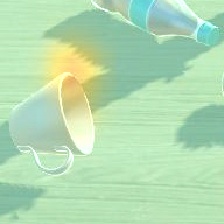}
  \includegraphics[width=0.3\columnwidth]{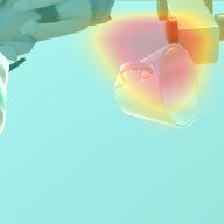} \vskip 1pt
\imgrow\label{fig:can}
  \includegraphics[width=0.3\columnwidth]{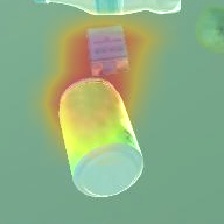}
  \includegraphics[width=0.3\columnwidth]{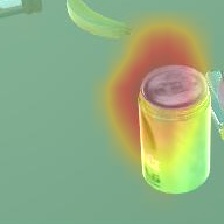}
  \includegraphics[width=0.3\columnwidth]{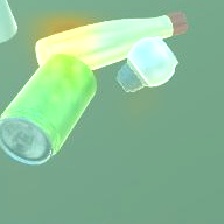} 
\end{imgrows}

\caption{Attended damaging collision regions (red) for different scene configurations with similar obstacles. Each row refers to a given obstacle  (a) Black tea pot, (b) White cup , (c) Green can.}
\label{fig:generalization}
\end{figure}

PonNet generalization ability was also assessed qualitatively from the samples. This analysis purpose is to determine whether PonNet simply detects objects (i.e., a given obstacle always predicted as DC or NDC) or PonNet is really able to predict the physical interactions between different objects, as initially intended. To this end, samples containing the same obstacles were evaluated. Some of these samples are illustrated in Fig. \ref{fig:generalization}  to support our claims. Each row refers to a given obstacle. For the sake of simplicity, we represent only  RGB attention maps. For this modality areas of higher risk of damaging collisions are attended. All these samples were correct (true-positive and true-negative).

In the first row, the tea pot is an obstacle that is highly likely to lead to DC. However, the middle sample is predicted as NDC: in this case there is enough space to place the target without colliding with the obstacle.
The second row, about the white cup, illustrates three different behaviors. In the left sample, DC is unlikely as the obstacle is upward, in equilibrium and there is enough space. In the two other samples, the obstacle is lying down. Although the middle sample is correctly predicted as NDC, the attention is slightly higher: the object was able to roll without any damaging collision. In the right sample, damaging collisions occurred,  because the cup rolled and hit the toy car. This  could be  predicted and captured by PonNet.
The third rows, shows a green can obstacle in different positions. When laying down, only the left sample is predicted DC as this obstacle is a central position where the target is supposed to be placed, which may lead to a strong collision. In the right sample case, this obstacle is slightly shifted and is able to roll when touched during the placing motion. When the can is upwards, it generates a DC as it is correctly predicted in the middle sample.

These results suggest that PonNet is able to generalize the DCP task and does not overfit to obstacles. Indeed, it can be inferred that physical properties of the obstacles are learned by the network. 

\subsection{Collision Type Analysis}
In a more extensive study, the different types of damaging collisions were also analyzed. Four additional types of damaging collisions were labeled from the simulator. The collision types can be described as follows:
\begin{itemize}
\item ({\bf AO}) Collision between the robot arm and an obstacle.
\item ({\bf TO}) Collision between the target and an obstacle.
\item ({\bf OO}) Collision between multiple obstacles.
\item ({\bf OD}) Collision between an obstacle and the destination.
\item ({\bf Any}) All types of damaging collisions.
\end{itemize}
It should be noted that for each sample one or several types of damaging collisions may occur. The results are given in Table~\ref{tab:collisions} for the average test set accuracy and standard deviation over five runs. In this set of experiments, we introduced the sum of each type of collision loss into the global loss function. Therefore, the accuracy of ({\bf Any}) collision was different from the results in Table \ref{tab:results}.  Please note that the plane detection method cannot solve such a task, and was therefore not tested for this set of experiments. 

 In comparison with the initial DCP task, predicting each type of collision is more complex. The accuracy of each of these collisions drastically decreased compared with the results obtained in Table \ref{tab:results}. In particular, it was relatively complex to predict damaging collisions related to ({\bf TO}). This is challenging because ({\bf TO}) collisions depend on the speed and the shape of the target in addition to the force of the robot motion. 
Although, ({\bf Any}) collisions are correlated with the other collisions, the difference in accuracy may be explained by the strong non-linearity between modalities, intra-branches (see Table \ref{tab:conf}) and between each collision branches, inter-branches. Furthermore, We assume that the different PonNet architectures and loss functions were not optimized to predict several types of collision at the same time. Indeed, We observed that the prediction accuracy of ({\bf Any}) collision, for all PonNet types, was decreased compared with our first set of experiments. Future studies about the architectures and the intra- and inter-branches relations may help to improve this kind of task.

\begin{table*}
\setlength\tabcolsep{2.8pt}
\caption{Average test set accuracy and standard deviation over five trials for different types of collisions considering the different ablations of PonNet. }
\label{tab:collisions}
\small
\begin{tabular}{l|c|c|c|c|c|c}
\hline
 & {\bf Input} &\multicolumn{5}{c}{\bf Collision type }  \\
\cline{3-7}
 \multicolumn{1}{l|}{\bf Method }  &{\bf  Type}& {\bf Any}& {\bf AO} &{\bf TO}&{\bf OO} & {\bf OD}\\
\hline 
 &RGB &87.74 $\pm$ 0.53 &79.00 $\pm$ 1.51 & 78.02 $\pm$ 0.65& 81.30 $\pm$ 0.73 & 80.42 $\pm$ 0.60\\
 PonNet-type1  &D & 86.38 $\pm$ 0.65 & 77.83 $\pm$ 0.70 &77.16 $\pm$ 0.72& 82.46 $\pm$ 1.01 & 80.21 $\pm$ 0.33\\
 &RGBD&88.05 $\pm$ 0.44 &79.22 $\pm$ 0.65 &77.96 $\pm$ 0.36& 81.14 $\pm$ 0.59 & 81.11 $\pm$ 0.94\\
 \hline
 &RGB &87.35 $\pm$ 1.29 &80.21 $\pm$ 0.52 &77.39 $\pm$ 0.59& 82.58 $\pm$ 1.56 & 81.96 $\pm$ 1.09\\
PonNet-type2 &D & 88.29 $\pm$ 0.60 & 80.93 $\pm$ 0.30 &77.19 $\pm$ 0.87& 82.67 $\pm$ 0.99 & 81.82 $\pm$ 0.53\\
 &RGBD&89.15 $\pm$ 0.46 &80.79 $\pm$ 1.68 &76.84 $\pm$ 1.49& 82.13 $\pm$ 0.21 & 82.10 $\pm$ 0.39\\
 \hline
 &RGB &88.94 $\pm$ 0.66 &79.50 $\pm$ 1.07 &76.44 $\pm$ 1.23& 81.43 $\pm$ 0.63 & 81.77 $\pm$ 0.79\\
 PonNet-type3  &D&87.93 $\pm$ 0.39 &77.48 $\pm$ 1.33 &74.01 $\pm$ 1.14& 82.02 $\pm$ 0.36 & 80.85 $\pm$ 1.81\\
 &RGBD&89.89 $\pm$ 0.76 &79.48 $\pm$ 0.95 &75.04 $\pm$ 1.60& 81.98 $\pm$ 0.74 & 82.13 $\pm$ 1.46\\
\hline
 PonNet-type4 &RGBD&89.90 $\pm$ 0.43 &79.18 $\pm$ 0.27 &76.96 $\pm$ 1.36& 81.96 $\pm$ 0.78 & 82.01 $\pm$ 0.89\\
 \hline
 PonNet&RGBD& 90.10 $\pm$ 0.33 &80.10 $\pm$ 0.12 & 77.25 $\pm$ 0.94& 83.01 $\pm$ 0.59& 82.11 $\pm$ 0.11\\
\hline
\end{tabular}
\end{table*}

\section{Conclusion}

Motivated by the increasing demand for DSRs, we developed a method to predict damaging collisions when DSRs are placing daily life objects in designated areas. In this study, we made the following contributions:
\begin{itemize}
\item We proposed PonNet which can predict damaging collisions from camera images obtained before placing motions. Unlike existing methods, PonNet has multiple attention branches and an additional self-attention mechanism to fuse the output of these branches.
\item We validated PonNet using a large-scale dataset for the DCP task. PonNet outperformed the baseline plane detection method by $8.4\%$. 
\end{itemize}
Future studies include combining this work with linguistic explanation.


\bibliographystyle{tfnlm}
\bibliography{refs}

\begin{thebibliography}{10}
\providecommand{\url}[1]{\normalfont{#1}}
\providecommand{\urlprefix}{Available from: }

\bibitem{smarr2014domestic}
Smarr~CA, Mitzner~T, et~al. {Domestic Robots for Older Adults: Attitudes,
  Preferences, and Potential}. {International Journal of Social Robotics}.
  2014;\hspace{0pt}6(2):229--247.

\bibitem{billard2019trends}
Billard~A, Kragic~D. Trends and challenges in robot manipulation. Science.
  2019;\hspace{0pt}364(6446).

\bibitem{brose2010role}
Brose~SW, Weber~DJ, et~al. {The Role of Assistive Robotics in the Lives of
  Persons with Disability}. American Journal of Physical Medicine \&
  Rehabilitation. 2010;\hspace{0pt}89(6):509--521.

\bibitem{gualtieri2018pick}
Gualtieri~M, ten Pas~A, Platt~R. {Pick-and-place without Geometric Object
  Models}. In: IEEE ICRA; 2018. p. 7433--7440.

\bibitem{magassouba2018multimodal}
Magassouba~A, Sugiura~K, Kawai~H. {A Multimodal Classifier Generative
  Adversarial Network for Carry-and-Place Tasks From Ambiguous Language
  Instructions}. IEEE RA-L. 2018;\hspace{0pt}3(4):3113--3120.

\bibitem{harada2014validating}
Harada~K, Tsuji~T, et~al. {Validating an Object Placement Planner for Robotic
  Pick-and-place Tasks}. Robotics and Autonomous Systems.
  2014;\hspace{0pt}62(10):1463--1477.

\bibitem{ABN_Fukui}
Fukui~H, Hirakawa~T, Yamashita~T, et~al. {Attention Branch Network: Learning of
  Attention Mechanism for Visual Explanation}. In: CVPR; 2019. p. 10705--10714.

\bibitem{zhou2016cvpr}
Zhou~B, Khosla~A, Lapedriza~A, et~al. {Learning Deep Features for
  Discriminative Localization}. In: CVPR; 2016.

\bibitem{watanabe2017survey}
Watanabe~T, Yamazaki~K, Yokokohji~Y. {Survey of Robotic Manipulation Studies
  Intending Practical Applications in Real Environments-object Recognition,
  Soft Robot Hand, and Challenge Program and Benchmarking}. Advanced Robotics.
  2017;\hspace{0pt}31(19-20):1114--1132.

\bibitem{magassouba2019understanding}
Magassouba~A, Sugiura~K, Trinh~Quoc~A, et~al. {Understanding Natural Language
  Instructions for Fetching Daily Objects Using GAN-Based Multimodal
  Target-Source Classification}. IEEE RA-L. 2019;\hspace{0pt}4(4):3884--3891.

\bibitem{jiang2012learning}
Jiang~Y, Lim~M, Zheng~C, et~al. {Learning to place new objects in a scene}.
  IJRR. 2012;\hspace{0pt}31(9):1021--1043.

\bibitem{zeng2018robotic}
Zeng~A, Song~S, et~al. {Robotic Pick-and-place of Novel Objects in Clutter with
  Multi-affordance Grasping and Cross-domain Image Matching}. In: IEEE ICRA;
  2018. p. 1--8.

\bibitem{mottaghi2016happens}
Mottaghi~R, Rastegari~M, Gupta~A, et~al. {“What happens if...” Learning to
  Predict the Effect of Forces in Images}. In: ECCV; Springer; 2016. p.
  269--285.

\bibitem{magassouba2019multi}
Magassouba~A, Sugiura~K, Kawai~H. {Multimodal Attention Branch Network for
  Perspective-Free Sentence Generation}. Conference on Robot Learning (CoRL).
  2019;\hspace{0pt}.

\bibitem{magassouba2020multimodal}
Magassouba~A, Sugiura~K, Kawai~H. {A Multimodal Target-Source Classifier With
  Attention Branches to Understand Ambiguous Instructions for Fetching Daily
  Objects}. IEEE RA-L. 2020;\hspace{0pt}5(2):532--539.

\bibitem{chen2018progressively}
Chen~H, Li~Y. {Progressively Complementarity-aware Fusion Network for RGB-D
  Salient Object Detection}. In: CVPR; 2018. p. 3051--3060.

\bibitem{yang2019adaptive}
Yang~J, Zhou~X, Zhou~Z, et~al. {Adaptive Fusion of RGBD Data for two-stream
  FCN-based Level Set Tracking}. In: IEEE VCIP; 2019. p. 1--4.

\bibitem{bakhtin2019phyre}
Bakhtin~A, van~der Maaten~L, Johnson~J, et~al. {PHYRE: A New Benchmark for
  Physical Reasoning}. In: Advances in Neural Information Processing Systems;
  2019. p. 5083--5094.

\bibitem{johnson2017clevr}
Johnson~J, Hariharan~B, van~der Maaten~L, et~al. {CLEVR: A Diagnostic Dataset
  for Compositional Language and Elementary Visual Reasoning}. In: CVPR; 2017.
  p. 2901--2910.

\bibitem{yamamoto2018human}
Yamamoto~T, Nishino~T, Kajima~H, et~al. {Human Support Robot (HSR)}. In: Acm
  siggraph; 2018. p. 1--2.

\bibitem{inamura2013development}
Inamura~T, Tan~JTC, Sugiura~K, et~al. {Development of Robocup@Home Simulation
  Towards Long-term Large Scale HRI}. In: Robot Soccer World Cup; Springer;
  2013. p. 672--680.

\bibitem{I15}
Iocchi~L, Holz~D, Ruiz-del Solar~J, et~al. {RoboCup@Home: Analysis and Results
  of Evolving Competitions for Domestic and Service Robots}. {Artificial
  Intelligence}. 2015;\hspace{0pt}229:258--281.

\bibitem{gupta2014learning}
Gupta~S, Girshick~R, Arbel{\'a}ez~P, et~al. {Learning Rich Features from RGB-D
  Images for Object Detection and Segmentation}. In: ECCV; Springer; 2014. p.
  345--360.

\bibitem{aakerberg2017depth}
Aakerberg~A, Nasrollahi~K, Rasmussen~CB, et~al. {Depth Value Pre-Processing for
  Accurate Transfer Learning based RGB-D Object Recognition}. In: IJCCI; 2017.
  p. 121--128.

\bibitem{hori2017attention}
Hori~C, Hori~T, et~al. {Attention-based Multimodal Fusion for Video
  Description}. In: ICCV; 2017. p. 4193--4202.

\bibitem{russakovsky2015imagenet}
Russakovsky~O, Deng~J, et~al. {ImageNet Large Scale Visual Recognition
  Challenge}. IJCV. 2015;\hspace{0pt}115(3):211--252.

\bibitem{he2016identity}
He~K, Zhang~X, Ren~S, et~al. {Identity Mappings in Deep Residual Networks}. In:
  ECCV; Springer; 2016. p. 630--645.

\bibitem{wang2018plane}
Wang~C, Guo~X. {Plane-Based Optimization of Geometry and Texture for RGB-D
  Reconstruction of Indoor Scenes}. In: IEEE 3DV; 2018. p. 533--541.

\end{thebibliography}

\end{document}